\documentclass[runningheads,a4paper]{llncs}

\usepackage{amssymb}
\usepackage{graphicx}
\usepackage[mathscr]{eucal}
\usepackage{amsmath}
\usepackage{amsfonts}

\usepackage{url}
\urldef{\mailsa}\path|hagi@edu.mie-u.ac.jp|
\newcommand{\keywords}[1]{\par\addvspace\baselineskip
\noindent\keywordname\enspace\ignorespaces#1}

\def\E{\mathbb{E}}
\def\x{{\boldsymbol{x}}}
\def\y{{\boldsymbol{y}}}
\def\0{{\bf 0}}
\def\I{{\bf I}}
\def\O{{\bf O}}
\def\X{{\bf X}}

\def\U{{\bf U}}

\def\y{{\boldsymbol{y}}}
\def\w{{\boldsymbol{w}}}

\def\r{{\boldsymbol{r}}}

\def\tp{{^{\top}}}
\def\evw{\widehat{\boldsymbol{w}}}
\def\vDelta{{\boldsymbol{\Delta}}}
\def\Ssse{S^{\rm SSE}}
\def\rsse{\r^{\rm SSE}}
\def\Smse{S^{\rm MSE}}
\def\rmse{{\boldsymbol{r}}^{\rm MSE}}
\def\Smb{S^{\rm MB}}

\def\Dsse{\vDelta^{\rm SSE}}
\def\Dmse{\vDelta^{\rm MSE}}
\def\Dmb{\vDelta^{\rm MB}}

\def\rmse{\r^{\rm MSE}}

\begin{document}

\mainmatter  

\title{On gradient descent training under data
augmentation with on-line noisy copies}

\titlerunning{Gradient descent under DA with on-line noisy copies}

%
%
\author{Katsuyuki Hagiwara}
\authorrunning{K. Hagiwara}

\institute{Faculty of Education, Mie University\\
1577 Kurima-Machiya-cho, Tsu, 514-8507, Japan\\
\mailsa}

%
%

\toctitle{Gradient descent under data augmentation with noisy copies}
\tocauthor{Gradient descent under data augmentation with noisy copies}
\maketitle

\begin{abstract}
In machine learning, data augmentation (DA) is a technique for improving
the generalization performance.  In this paper, we mainly considered
gradient descent of linear regression under DA using noisy copies of
datasets, in which noise is injected into inputs.  We analyzed the
situation where random noisy copies are newly generated and used at each
epoch; i.e., the case of using on-line noisy copies. Therefore, it is
viewed as an analysis on a method using noise injection into training
process by DA manner; i.e., on-line version of DA. We considered the
training process under three training situation which are the full-batch
training under the sum of squared errors, the full-batch and mini-batch
training under the mean squared error.  We showed that, in all cases,
training for DA with on-line copies is approximately equivalent to the
$\ell_2$ regularization training whose regularization parameter
corresponds to the variance of injected noise.  On the other hand, we
showed that DA with on-line copies yields apparent acceleration of
training in full-batch under the sum of squared errors and the
mini-batch under the mean squared error; i.e., the learning rate is
multiplied by the number of noisy copies plus one. The apparent
acceleration and regularization effect come from the original part and
noise in a copy data respectively.  These results are confirmed in a
numerical experiment.  In the numerical experiment, we found that our
result can be applied to usual off-line DA in under-parameterization
scenario and can not in over-parametrization scenario. Moreover, we
experimentally investigated the training process of neural networks
under DA with off-line noisy copies and found that our analysis on
linear regression is possible to be applied to neural networks.
\end{abstract}

\section{Introduction}

\subsection{Background}

Data augmentation (DA) is a technique for increasing the number of data
and applied in many deep learning applications; e.g., image
\cite{Shorten2019} including major convolutional neural network
architectures such as \cite{AlexNet} and successors,
speech\cite{Ramirez2019} and natural language processing\cite{Li2022}
and so on.  Despite of many reports showing the effectiveness of DA, it
is not so clear how and why it works.

Generally, DA is known as a strategy for avoiding over-fitting and
improving robustness. It is brought about by increasing the number of
data which is meaningful in some sense.  In image tasks, augmented data
are often made by geometric transformation of the original data such as
rotation, scale and position changes and so on. DA by these processings
may append the new data sets which are possible to appear in prediction
phase.  In an another viewpoint, we may be able to say that this
provides a robustness in terms of change of view of objects.  This
effect seems to be bias reduction in the generalization error.  On the
other hand, there is a DA technique that appends new data sets whose
inputs are disturbed by noise.  This is possible to keep robustness in
terms of input changes in some sense; e.g., injection of noise into
training inputs to overcome adversarial examples \cite{Panda2021}. In
addition to the rboustness, it is often pointed out that this type of DA
provides a regularization effect by referring
\cite{Bishop1995}. Therefore, it may reduce variance in the
generalization error. Note that it is not clear the distinction of
effect of popular DA techniques using interpolation of data such as
\cite{NFM}. This paper concerns with the DA technique using noise
injection into inputs.

\subsection{Contribution}

In this paper, we study a naive DA technique using noise injection into
input data. More precisely, a set of pairs of input disturbed by noise
and output for the original input are appended to the original data
set. We call this augmented data set {\em a noisy copy} of the original
data set; i.e., if the number of data is $n$ and the number of copy sets
is $K$, the number of whole data under augmentation is $(K+1)n$, where
$1$ for the original data.  Obviously, this increases the number of data
as desired. Note that noise is injected into only input data here while
there is a method for disturbing also labels\cite{NoisyLabels}. We focus
on a training process under DA using a noisy copies. We consider
gradient descent (GD) of linear regression. Although this is a quite
simple setting, it is often assumed in the analysis related to neural
networks and learning, especially in over-parametrization scenario;
e.g., \cite{Hastie2020,NTK}. It is also related to the analysis of
randomized neural network; e.g. \cite{RandNet}. Unfortunately, it is not
straightforward to analyze the GD training process under the above
setting of DA even when assuming linear regression.  As argued later
precisely, this is because weight update at any epoch is correlated to
injected noise since noise is injected before training and fixed in
training.  We refer to this type of noisy copy as {\em off-line} noisy
copy.  Note that this is a natural setting in the context of DA.  In
this paper, we consider the DA with {\em on-line} noisy copy which is
randomly drawn and injected at each epoch in training. Indeed, this can
be viewed as a kind of noise injection in training process as argued
later. Therefore, from another point of view, we analyze a method using
noise injection into training process by DA manner; i.e., on-line
version of DA.

In this paper, we analyzed the GD update of linear regression under DA
with random on-line noisy copies.  The main contribution of this paper
is as follows.
\begin{itemize}
 \item For the full-batch training under the sum of squared errors, we
       showed that the GD update at an epoch is approximately equivalent
       to that of the $\ell_2$ regularization (ridge regression) with
       $K+1$ times learning rate, where $K$ is the number of copies.

 \item For the full-batch training under the mean squared error, we
       showed that the GD update at an epoch is approximately equivalent
       to that of the $\ell_2$ regularization.

\item For the mini-batch training under the mean squared error, by
       ignoring the update term including the square of the learning
       rate, we showed that the GD update at an epoch is approximately
       equivalent to that of the $\ell_2$ regularization with $K+1$ times
       learning rate.
\end{itemize}
These results rely on the evaluation based on the expected value in
three cases and, additionally, almost surely convergence for the second
case. Note that the results here do not depend on whether the scenario is
under-parametrization or over-parametrization. In this paper, these
results were confirmed in a simple numerical experiment.

In all results, the regularization parameter is proportional to the
variance of noise; i.e., the smoothing effect is high when the noise
variance is large. This result is intuitively natural while it is
explicitly formulated here due to the analysis of a simple linear regression.

On the other hand, apparent acceleration is found in the full-batch
training under the sum of squared errors and the mini-batch
training. This seems to be natural because, in DA, similar data sets are
presented several times in one epoch.  However, this is not
straightforward since it can not be observed in the full-batch under the
mean squared error. Especially, a simple and interpretable result in the
mini-batch training is approximately obtained and may not be trivial, in
which approximation is valid when the learning rate is small or the
update term is small.

In our result, apparent acceleration applies also to DA with
off-line noisy copies, which is natural situation of DA. On the one hand,
the regularization effect of off-line case can not be specified while it
exists as shown in our numerical experiment. It is stochastically determined by
noise injected before training.  However, the numerical experiment
revealed that the regularization effects for both type of DA are very
similar in under-parametrization scenario while they are not in
over-parametrization scenario.  This is because, in the
over-parameterization scenario, over-fitting to noise injected before
training is serious; i.e., the training process depends largely on the
injected noise.

The important point in our result for linear regression is that there is
nothing but apparent acceleration and regularization in the effect of
DA.  And, the regularization effect by DA is important for
generalization while the increase in the total number of data by DA may
not so important other than apparent acceleration of training.  It is
worthwhile to note that the original input data and noise in copy
contribute to apparent acceleration and effect of reguralization respectively.

Furthermore, for layered neural networks, we execute a simple numerical
experiment of the full-batch and mini-batch GD training under the mean
squared error criterion. We compared the training curves with off-line
DA and without DA for each training condition.  The results are
summarized below.
\begin{itemize}
 \item In the full-batch training, apparent acceleration is not observed
       when introducing DA while the effect of regularization is
       observed.
\item In the mini-batch training, both of 
apparent acceleration and effect of regularization are observed.
\end{itemize}
The training process of neural networks was very similar to that of
linear regression. This may imply that our findings for linear
regression are possible to apply to neural networks.  Therefore, we need
to take this into account in applying DA with off-line noisy copies in
the deep learning software such as Keras\cite{Keras}. On the other hand,
it is worthwhile to note that a mixture of apparent acceleration and
regularization effect by DA may has a different effect on training
process of neural networks. This is because there are bottlenecks in
training of neural networks; e.g. \cite{Dinh2017}.  Unlike linear
regression, the error surface of neural networks is complex.  Because of
this, there may exist the effects that are not appeared in the analysis
of linear regression; i.e., the other than a simple acceleration and
regularization. The investigation of this is left as a future work.

\section{Related works}

As a work on DA with off-line noisy copies, noise injection into the
input is shown to be equivalent to the training under a cost function
with a regularization term\cite{Bishop1995}. It is not argued the
introduction of noisy copies and the training process.  \cite{Zur2009}
experimentally showed that, in classification tasks, input noise
injection can reduce over-fitting compared to the early stopping and as
weight decay.  However, there is no analysis on the training process
under DA with noisy copies.  On the other hand, we considered DA with
on-line noisy copies.  Indeed, this can be viewed as a kind of noise
injection in training process, in which noise is injected into inputs,
weights and hidden output;
e.g. \cite{Dropout,Ho2008,Camuto2020,Dhifallah2021,Panda2021}.  In this
direction, there are a lot of works to overcome adversarial examples
especially in images; e.g., \cite{Panda2021}.  Among these,
\cite{Dhifallah2021} is the most relevant work that provided an
asymptotic characterization of the training and generalization errors of
input noise injection under a random feature model.  This is more
general than our result since it considered a nonlinear model while it
is not considered a difference in training criterion. Also, our result
is more simple and interpretable due to the restriction to a linear
regression problem. Nevertheless, it may be meaningful for understanding
the training process of neural networks since it includes an
over-parameterized case; e.g., \cite{NTK}. Actually, our experimental
result showed that the analysis on linear regression seems to be
applicable to the training of neural networks.  Despite of our simple
setting of linear regression, we can not find similar works to
ours. This is because the DA based on-line input noise injection may
be minor in practices.

\section{Gradient descent of linear regression}

We consider a linear regression problem.  Let $\y$ and $\X$ be an
$n$-dimensional response (output) vector and an $n\times m$ matrix of
predictors (inputs).  $n$ is the number of data and $m$ is the number of
input variables (input dimension).  The $i$th element of $\y$ is denoted
by $y_i$ and the $(i,j)$ element of $\X$ is denoted by $x_{i,j}$;
i.e. the $i$th data of the $j$th variable.  In the context of DA below,
$(\X,\y)$ is referred to as the original data.  Let $\w$ is an
$m$-dimensional weight (coefficient) vector, whose $j$th element is
denoted by $w_j$.

We consider gradient descent (GD) under an error function
(criterion) on $\w$, in which we assume that the loss function is squared
error.  We define
\begin{align}
\label{eq:SSE-for-X-y}
S(\w)&:=\|\y-\X\w\|^2,
\end{align}
which is the sum of squared errors (SSE).  For example, GD according to
the SSE is an algorithm, in which, by starting from
an initial vector $\w(0)$, the $t$th step (epoch) of the update is given
by
\begin{align}
\w(t)&=\w(t-1)-\eta\vDelta_S(t-1)\\
\label{eq:Delta_S-t-1}
\vDelta_S(t-1)&:=\left.\frac{\partial S(\w)}{\partial \w}\right|_{\w=\w(t-1)}
=-2\X\tp(\y-\X\w(t-1)),
\end{align}
where $\eta>0$ is a constant step size or learning rate.  Note that
training according to $S(\w)$ is the full-batch (FB) training on data
$(\X,\y)$.  If the error function is the $\ell_2$ regularized cost
function with SSE defined by
\begin{align}
\label{eq:regcost_SSE}
C(\w)&:=S(\w)+\lambda\|\w\|^2,
\end{align}
the update term is given by
\begin{align}
\label{eq:vDelta_C}
\vDelta_C(t-1)&:=\vDelta_S(t-1)+2\lambda\w(t-1).
\end{align}

\section{Data augmentation with noisy copies}

\subsection{Data augmentation}

We now consider a data augmentation (DA) by adding the $K$ pairs of noisy
copies of input matrix $\X$ and the corresponding output $\y$.

Let $k$ and $t$ be indexes of noisy copy and epoch (step) in training,
where $k\in\{1,\ldots,K\}$ and $t\in\{0,1,\cdots\}$.  L et $\U_{t,k}$ be
an $n\times m$ random matrix whose elements are samples from
$N(0,\tau^2/n)$, where $\tau >0$ and $K$ is the number of copies.  We
define $\U_{t,0}=\O_{n,m}$, where $\O_{n,m}$ is an $n\times m$ zero
matrix.  We then define
\begin{align}
\X_{K,t}=
\begin{bmatrix}
\X+\U_{t,0}\\
\X+\U_{t,1}\\
\vdots\\
\X+\U_{t,K}
\end{bmatrix},~
\y_K=
\begin{bmatrix}
\y\\
\vdots\\
\y
\end{bmatrix},
\end{align}
where $\y_K$ consists of a $K+1$ times repetition of $\y$.  We also
define $\y_0=\y$ and $\X_{0,t}=\X$. 
We refer to $(\X_{0,t},\y_0)=(\X,\y)$ as the original data.
We assume that samples in
$\{\U_{t,k}\mid 1\le k\le K,~0\le t\}$ are mutually independent. For
simplicity, we assume that $(\X,\y)$ and the initial value of the weight
vector are non-stochastic.  We focus on the stochastic behavior
originated from $\U_{t,k}$'s.

We explicitly refer to $\U_{t,k}$ as on-line noise. On the other hand,
we write $\U_k=\U_{t,k}$ if noisy copies are unchanged during
training. This may be usual case of DA.  In this case, we refer to
$\U_k$ as off-line noise.

\subsection{FB training under SSE}

For a weight vector $\w$, we define SSE for augmented data by
\begin{align}
\Ssse_{K,t}(\w):=\|\y_K-\X_{K,t}\w\|^2,
\end{align}
where we set $\Ssse_{0,t}(\w):=S(\w)$; i.e., $\Ssse_{0,t}$ is the SSE
for the original data, which is defined in (\ref{eq:SSE-for-X-y}). The
GD training under $\Ssse_{K,t}$ is a FB training and the GD step is
given by
\begin{align}
\w(t)&=\w(t-1)-\eta\Dsse_K(t-1)\\
\Dsse_K(t-1)
&:=\left.\frac{\partial\Ssse_{K,t}(\w)}{\partial \w}\right|_{\w=\w(t-1)}.
 \end{align}
Since
\begin{align}
\label{eq:Ssse_K}
\Ssse_{K,t}(\w)&=\|\y-\X\w\|^2+\sum_{k=1}^K\|\y-(\X+\U_{t,k})\w\|^2\notag\\
&=S(\w)+\sum_{k=1}^K\left\{
\|\y-\X\w\|^2+\|\U_{t,k}\w\|^2+2(\y-\X\w)\tp\U_{t,k}\w\right\}\notag\\
&=(K+1)S(\w)+\sum_{k=1}^K
\left\{\|\U_{t,k}\w\|^2+2\y\tp\U_{t,k}\w-2\w\tp\X\tp\U_{t,k}\w\right\}
\end{align}
holds, by the definition of $S$, we obtain
\begin{align}
\label{eq:update-term-S_K}
\Dsse_K(t-1)&=(K+1)\left[\vDelta_{S}(t-1)+\frac{2}{K+1}\rsse_K(t-1)\right]\\
\label{eq:r_K-in-update-term}
\rsse_K(t-1)&:=\sum_{k=1}^K\left\{
\U_{t,k}\tp\y-(\X\tp\U_{t,k}+\U_{t,k}\tp\X-\U_{t,k}\tp\U_{t,k})\w(t-1)\right\},
\end{align}
where $\Delta_S$ is defined in (\ref{eq:Delta_S-t-1}).  

Regardless of on-line or off-line noise, as found in
(\ref{eq:update-term-S_K}), the update term under $\Ssse_{K,t}$ is $K+1$
times larger than that under $S$ which is SSE for the original data.
Since $\Dsse_K$ is multiplied by the learning rate, the effect of adding
noisy copies to the training data can be viewed as the use of
$(K+1)\eta$; i.e., the role of increasing the learning rate. Obviously,
this can be avoided by normalizing $\U_{t,k}$. We discuss this later.

On the other hand, the behavior of $\rsse_K(t-1)$ may not so clear.  We
consider the case of $m\le n$. Let $\evw$ and $L(\X)$ be the least
squares estimator under the original data and the linear subspace
spanned by the column vectors of $\X$.  $\X\evw$ is given by the
orthogonal projection of $\y$ onto $L(\X)$.  Then, in the GD training
under $S$, $\w(t)$ is updated so that $\X\w(t)$ moves toward the
$\X\evw$ on $L(\X)$.  In case of injecting noise, this happens on
$L(\X+\U_{t,k})$. In other words, injection of $\U_{t,k}$ disturbs the
convergence to the least squares estimate on $L(\X)$.  However,
generally, we cannot specify whether it improves the generalization
performance or not; i.e., it depends on the linear subspace generated by
$\X+\U_{t,k}$.

We consider the influence of $\rsse_K(t-1)$.  If $\U_{t,k}=\U_{k}$ then
$\rsse_K(t-1)$ is deterministic during the training; i.e., in case of
off-line noise. In this case, it is difficult to analyze the
probabilistic behavior of this term since $\U_{k}$ is correlated with
$\w(t-1)$.  However, if $\U_{t,k}$ is randomly drawn at each step,
$\U_{t,k}$ is independent of $\w(t-1)$. This is because $\w(t-1)$ is a
function of $\{\U_{k,s}\mid 1\le k\le K,~0\le s< t\}$ which is
independent of $\U_{t,k}$, $k=1,\ldots,K$. In this case, by
(\ref{eq:r_K-in-update-term}), we obtain
\begin{align}
\E_{\U_{t,k}}\left[\rsse_K(t-1)\right]&=K\tau^2\w(t-1)
\end{align}
since the elements of $\U_{t,k}$ are i.i.d. according to $N(0,\tau^2/n)$, where
$\E_{\U_{t,k}}$ is the expectation with respect to the joint
probability distribution of the elements of $\U_{t,k}$.
Therefore, we obtain
\begin{align}
\E_{\U_{t,k}}\left[\Dsse_K(t-1)\right]
&=(K+1)\left[\vDelta_S(t-1)+2\frac{K\tau^2}{K+1}\w(t-1)\right].
\end{align}
The update rule according to $\Ssse_{t,k}$ is essentially equivalent to
the rule under the $\ell_2$ regularized cost which is given by
(\ref{eq:regcost_SSE}), in which the regularization parameter is
$\lambda=K\tau^2/(K+1)$. However, there is a difference in learning
rate. In the update under $\Ssse_{t,k}$, it is automatically multiplied
by $(K+1)$; i.e., the learning rate depends on the number of copies. It
implies that apparent acceleration occurs and the effect is notable when
the number of copies is large.

Since we present a similar data several times at one epoch in the DA
setting, apparent acceleration is intuitively understood.  However, it is
not straightforward as found in the next analysis.

\subsection{FB training under mean squared errors}

We next consider to employ the mean squared error (MSE) instead of the
SSE above. 
It is defined by
\begin{align}
\Smse_{t,k}(\w):=\frac{1}{(K+1)n}\Ssse_{t,k}(\w),
\end{align}
since the number of data is $(K+1)n$ in this case.  We define
$\Smse_{0,t}(\w)=\frac{1}{n}S(\w)$ that is the MSE for the original
data.  Note that this is FB training. The GD step for $\Smse_K$ is given by
\begin{align}
\w(t)&=\w(t-1)-\eta\Dmse_K(t-1)\\
\Dmse_K(t-1)
&:=\left.\frac{\partial \Smse_{t,k}(\w)}{\partial \w}\right|_{\w=\w(t-1)}.
 \end{align}
By (\ref{eq:Ssse_K}), we obtain
\begin{align}
\Smse_{t,k}(\w)&=\frac{1}{n}S(\w)+\frac{1}{n}\frac{1}{K+1}\sum_{k=1}^K
\left\{\|\U_{t,k}\w\|^2+2\y\tp\U_{t,k}\w-2\w\tp\X\tp\U_{t,k}\w\right\}.
\end{align}
and
\begin{align}
\label{eq:update-term-S_K-mse}
\Dmse_K(t-1)
&=\frac{1}{n}\vDelta_S(t-1)+\frac{2}{n}\frac{K}{K+1}\rmse_K(t-1)\\
\label{eq:r_K-in-update-term-mse}
\rmse_K(t-1)&:=\frac{1}{K}\sum_{k=1}^K\left\{
\U_{t,k}\tp\y-(\X\tp\U_{t,k}+\U_{t,k}\tp\X-\U_{t,k}\tp\U_{t,k})\w(t-1)\right\}.
\end{align}
By the same story as in case of the SSE, we obtain
\begin{align}
\E_{\U_{t,k}}\left[\rmse_K(t-1)\right]&=\tau^2\w(t-1)
\end{align}
and, therefore, 
\begin{align}
\label{eq:E-Delta-Smse_K}
\E_{\U_{t,k}}\left[\Dmse_K(t-1)\right]
&=\frac{1}{n}\vDelta_S(t-1)+2\frac{\tau^2}{n}\frac{K}{K+1}\w(t-1).
\end{align}
The update under $\Smse_K$ is essentially equivalent to the update under
the $\ell_2$ regularized cost which is given by
$S(\w)/n+\lambda\|\w\|^2$, in which the regularization parameter is
$\lambda=\tau^2K/(K+1)n$. In this case, the learning rate is not
affected by $K$.  This is because of the normalization by $K+1$ in
$\Smse_{t,k}$.

In this case, the other type of the probabilistic behavior makes sense
as shown in \cite{Dhifallah2021} .  We fix $n$ and
$m$.  Note that all elements of the matrix in $\{\U_{t,k}\mid
k=1,\ldots,K\}$ are mutually independent and those are also mutually
independent of the elements of $\y$, $\X$ and $\w(0)$.  Let
$u_{t,k}^{i,j}$ be the $(i,j)$ element of $\U_{t,k}$. Since
$u_{t,k}^{i,j}$, $k=1,\ldots,K$ are i.i.d. according to $N(0,\tau^2/n)$,
we obtain $\lim_{K\to\infty}\frac{1}{K}\sum_{k=1}^Ku_{t,k}^{i,j}=0$
almost surely.  Therefore,
$\lim_{K\to\infty}\sum_{k=1}^K\U_{t,k}=\O_{n,m}$ almost surely. On the
other hand, we define
$Z_{t,k,j_1,j_2}=\sum_{i=1}^nu_{t,k}^{i,j_1}u_{t,k}^{i,j_2}$.  Note that
the elements in $\U_{t,k}$ are i.i.d according to $N(0,\tau^2)$ for any
$k$ and $t$.  Therefore, for any $j_1$ and $j_2$, $Z_{t,k,j_1,j_2}$,
$k=1,\ldots,K$ are i.i.d with $\E\left[Z_{t,k,j_1,j_2}\right]=0$ if
$j_1\neq j_2$ and with $\E\left[Z_{t,k,j_1,j_2}\right]=\tau^2$ if
$j_1=j_2$.  Therefore,
$\lim_{K\to\infty}\sum_{k=1}^K\U_{t,k}\tp\U_{t,k}=\tau^2\I_{m}$ almost
surely, where $\I_m$ is the $m\times m$ identity matrix. As a result, we
obtain
\begin{align}
\lim_{K\to\infty}\rmse_K(t-1)=\tau^2\w(t-1)
\end{align}
almost surely and, therefore, 
\begin{align}
\lim_{K\to\infty}\Dmse_K(t-1)&=\frac{1}{n}\vDelta_S(t-1)+2\frac{\tau^2}{n}\w(t-1)
\end{align}
almost surely. By (\ref{eq:E-Delta-Smse_K}), this is consistent with
$\E_{\U_{t,k}}\left[\Dmse_K(t-1)\right]$ for a large $K$.  In
\cite{Dhifallah2021}, it appears a different regularization term that
comes from a nonlinearity of random feature model.

This result on MSE is important. Since we present a similar data several times
at one epoch in the DA setting, the training speed under DA seems to be
high compared to presenting only the original data. However, the above
result implies that there is no apparent acceleration effect under the
MSE setting. This result, together with the result for the previous SSE
setting, it may depends on the training criterion. The next question is
what happens in the mini-batch (MB) setting which is a standard setting in
deep learning.

\subsection{MB training unser MSE}

We first consider the mini-batch (MB) training for the original data.

Let $\rho$ be the MB size and $Q$ be the number of MB, where $0<\rho<
n$.  We assume that the number of the original data is divisible by
$\rho$; i.e., $\rho=n/Q$.  We then set
$B_q=\{(q-1)\rho+1,\ldots,q\rho\}$ for $q=1,\ldots,Q$.  
$\{B_1,\ldots,B_Q\}$ be a partition of $\{1,\ldots,n\}$; i.e.,
$\{1,\ldots,n\}=\bigcup_{q=}^QB_q$ and $B_{q_1}\bigcap
B_{q_2}=\varnothing$ for $q_1\neq q_2$.  $B_q$ is the set of indexes of
the $q$th MB of the original data.

Let $\x_i$ be the $i$th row vector of $\X$.  Let $\X[B_q]$ be a
$\rho\times m$ input sub matrix whose row vectors are $\{\x_i\mid i\in
B_q\}$ with keeping the order.  Let $\y[B_q]$ be an output vector whose
elements are $\{y_i\mid i\in B_q\}$ with keeping the order. We define
\begin{align}
\label{eq:Smb_q}
\Smb_q(\w):=\frac{1}{\rho}\|\y[B_q]-\X[B_q]\w\|^2
\end{align}
which is the MSE for the $q$th MB. By setting
$\w(t,0)=\w(t-1)$, 
the
update in the MB training at $t$th epoch is that we repeat
\begin{align}
\label{eq:mb0-w-t-q}
\w(t,q) &=\w(t,q-1)-\eta\Dmb_q(t)\\
\label{eq:mb0-D-q}
\Dmb_q(t)
&:=\left.\frac{\partial\Smb_q(\w)}{\partial \w}\right|_{\w=\w(t,q-1)}\\
\label{eq:mb0-D-q-form}
&=-2\X[B_q]\tp(\y[B_q]-\X[B_q]\w(t,q-1))
\end{align}
for $q=1,\ldots,Q$ and we obtain $\w(t)=\w(t,Q)$.
We define
\begin{align}
\label{eq:mb0-D-a:b}
\Dmb_{a:b}(t)&:=\sum_{j=a}^{b}\Dmb_j(t).
\end{align}
Then, this process can also be written as
\begin{align}
\label{eq:mb0-w-t}
\w(t)&=\w(t-1)-\eta\Dmb_{1:Q}(t).
\end{align}

We next consider to apply the MB training for the whole data including
noisy copies. To achieve this, we may set
$A_p=\{(p-1)\rho,\ldots,p\rho\}$ for $p=1,\ldots,P_K$, where
$P_K=Q(K+1)$. Roughly speaking, at the $t$th epoch, we repeat the update
by starting from $\w(t-1)$ and choosing $(\X_{t,k}[A_p],\y_K[A_p])$ as
the MB. This sequential construction of the MB can also be written by
the double-loop in which the outer loop is for $k$ and inner loop is for
the MB.  We describe this procedure below.

For a weight vector $\w$, we define the error function in the MB training by
\begin{align}
\label{eq:Smbkq}
\Smb_{k,q}(\w):=\frac{1}{\rho}\|\y[B_q]-(\X[B_q]+\U_{t,k}[B_q])\w\|^2,
\end{align}
where $\Smb_{0,q}(\w)=\Smb_{q}(\w)$.  We define the update rule of the
MB training under DA as follows. For a fixed $k$, we define
\begin{align}
\label{eq:v-t-k-q}
\w(t,k,q) &=\w(t,k,q-1)-\eta\Dmb_{k,q}(t)\\
\label{eq:Delta-Smbkq-t-k-q-1}
\Dmb_{k,q}(t)
&:=\left.\frac{\partial\Smb_{k,q}(\w)}{\partial \w}\right|_{\w=\w(t,k,q-1)}
\end{align}
for $q=1,\ldots,Q$, where $\w(t,k,0)=\w(t,k-1,Q)$.  We
define $\w(t,k):=\w(t,k,Q)$. Now, in this algorithm, by setting
$\w(t,0,0)=\w(t-1)$ and repeating (\ref{eq:v-t-k-q}) with
(\ref{eq:Delta-Smbkq-t-k-q-1}) at each $k$, we obtain $\w(t,k)$ for
$k=0,1,\ldots,K$ successively. After all, we set $\w(t)=\w(t,K)$ which
is a resulting weight vector at epoch $t$. To clarify this algorithm,
the pseudo code is given in Fig.\ref{fig:code-mini-batch}.

\begin{figure}[h]
\begin{center}
\fbox{
\begin{minipage}{110mm}
\begin{enumerate}
\itemsep 0mm
\leftskip 0mm
\item $\w(t,0,0)=\w(t-1)$
\item for ($k\in\{0,1,\ldots,K\}$) \{
\item \qquad for ($q\in \{1,\ldots,Q\}$) \{
\item \qquad\qquad Obtain $\w(t,k,q)$ by (\ref{eq:v-t-k-q}) with (\ref{eq:Delta-Smbkq-t-k-q-1})
\item \qquad \}
\item \qquad $\w(t,k)=\w(t,k,Q)$
\item \}
\item $\w(t)=\w(t,K)$
\end{enumerate}
\end{minipage}
}
\end{center}
\caption{Code for the MB training}
\label{fig:code-mini-batch}
\end{figure}

For simplicity, we write $\X[B_q]=\X_q$, $\U_{t,k}[B_q]=\U_{t,k,q}$
and $\y[B_q]=\y_q$ below. Since 
\begin{align}
\frac{\partial\Smb_{k,q}(\w)}{\partial \w}
=&-\frac{2}{\rho}(\X_q+\U_{t,k,q})\tp(\y_q-(\X_q+\U_{t,k,q})\w)\notag\\
=&-\frac{2}{\rho}\X_q\tp(\y_q-\X_q\w)-\frac{2}{\rho}\U_{t,k,q}\tp(\y_q-\X_q\w)\notag\\
&+\frac{2}{\rho}\X_q\tp\U_{t,kq}\w+\frac{2}{\rho}\U_{t,k,q}\tp\U_{t,k,q}\w
\end{align}
holds, we obtain
\begin{align}
\label{eq:Dmb-0-q-t}
\Dmb_{0,q}(t)=-\frac{2}{\rho}\X_q\tp(\y_q-\X_q\w(t,0,q-1))
\end{align}
for $k=0$ and
\begin{align}
\label{eq:E_U-Dmb-k-q-t}
\E_{\U_{t,k,q}}\left[\Dmb_{k,q}(t)\right]
=&-\frac{2}{\rho}\X_q\tp(\y_q-\X_q\w(t,k,q-1))+\frac{2\tau^2}{n}\w(t,k,q-1).
\end{align}
for $k\ge 1$ since $\U_{t,k,q}$ and $\w(t,k,q-1)$ are mutually independent.
We use this fact for simplifying the description below.

We define
\begin{align}
 \Dmb_{k,a:b}(t):=\sum_{j=a}^b\Dmb_{k,j}(t).
\end{align}
When $k=0$, by the definition of DA here, the update
under the augmented data is the same as that under the original (noiseless) data.
We then have
\begin{align}
\label{eq:mb-v-t-0-q-suc}
\w(t,0,q)&=\w(t,0,q-1)-\eta\Dmb_{0,q}(t)
\end{align}
by (\ref{eq:v-t-k-q}), where $\w(t,0,0)=\w(t-1)$ . By this
relationship, we obtain
\begin{align}
\label{eq:mb0-v-t-0-Q-v-t-0-q}
 \w(t,0,Q)=\w(t,0,q)-\eta\Dmb_{0,q+1:Q}
\end{align}
and we then set $\w(t,1,0)=\w(t,0,Q)$.

For $k\ge 1$, by setting $\w(t,k,0)=\w(t,k-1,Q)$, we obtain
\begin{align}
\label{eq:v-t-k-q-v-t-k-1-Q}
\w(t,k,q)=\w(t,k-1,Q)-\eta\Dmb_{k,1:q} (t)
\end{align}
by (\ref{eq:v-t-k-q}) again.
We also obtain
\begin{align}
\label{eq:v-t-k-1-Q-v-t-k-1-0}
\w(t,k-1,Q)=\w(t,k-1,0)-\eta\Dmb_{k-1,1:Q} (t),
\end{align}
where $\w(t,k-1,0)=\w(t,k-2,Q)$ for $k\ge 2$ and $\w(k,0,0)=\w(t-1)$ for
$k=1$.  As a result, we obtain
\begin{align}
\label{eq:mb-v-t-k-q-v-t-0-q}
\w(t,k,q)&=\w(t,k-1,Q)-\eta\Dmb_{k,1:q}(t)
~({\rm by}~(\ref{eq:v-t-k-q-v-t-k-1-Q}))\notag\\
&=\w(t,k-1,0)-\eta\Dmb_{k-1,1:Q}-\eta\Dmb_{k,1:q}(t)
~({\rm by}~(\ref{eq:v-t-k-1-Q-v-t-k-1-0}))\notag\\
&=\w(t,k-2,Q)-\eta\Dmb_{k-1,1:Q}-\eta\Dmb_{k,1:q}(t)\notag\\
&=\w(t,k-2,0)-\eta\Dmb_{k-2,1:Q}-\eta\Dmb_{k-1,1:Q}-\eta\Dmb_{k,1:q}(t)
~({\rm by}~(\ref{eq:v-t-k-1-Q-v-t-k-1-0}))\notag\\
&\cdots\notag\\
&=\w(t,1,0)-\eta\sum_{j=1}^{k-1}\Dmb_{j,1:Q}-\eta\Dmb_{k,1:q}(t)
~({\rm by~ repeating~(\ref{eq:v-t-k-1-Q-v-t-k-1-0})})\notag\\
&=\w(t,0,Q)-\eta\sum_{j=1}^{k-1}\Dmb_{j,1:Q}-\eta\Dmb_{k,1:q}(t)\notag\\
&=\w(t,0,q)-\eta\Dmb_{q+1:Q}-\eta\Dmb_{k,1:q}(t)-
\eta\sum_{j=1}^{k-1}\Dmb_{j,1:Q},
\end{align}
where the last line comes from (\ref{eq:mb0-v-t-0-Q-v-t-0-q}).
This equation links the $q$th update of the $k$th copy to the $q$th
update of the original data at epoch $t$.
Therefore, we obtain
\begin{align}
\label{eq:mb-v-t-k-q-v-t-0-q-Ceta}
\w(t,k,q)=\w(t,0,q)-C\eta
\end{align}
by focusing on $\eta$, where $C$ is a constant that does not include
$\eta$ but, of course, is related to the other factors. On the other hand,
the update equation at $(t,k,q)$ is given by
\begin{align}
\label{eq:mb-v-t-k-q-v-t-k-q-1}
\w(t,k,q)=\w(t,k,q-1)-\eta\Dmb_{k,q}(t),
\end{align}
where $\w(t,k,0)=\w(t,k-1,Q)$. 
By (\ref{eq:E_U-Dmb-k-q-t}) and (\ref{eq:mb-v-t-k-q-v-t-0-q-Ceta}), we obtain
\begin{align}
&\E_{\U_{t,k,q}}\left[\Dmb_{k,q}(t)\right]\notag\\
=&\frac{1}{\rho}\left\{-2\X_q\tp(\y_q-\X_q\w(t,k,q-1))\right\}
+\frac{2\tau^2}{n}\w(t,k,q-1)\notag\\
=&\frac{1}{\rho}\left\{-2\X_q\tp(\y_q-\X_q\w(t,0,q-1))\right\}
+\frac{2\tau^2}{n}\w(t,0,q-1)+C'\eta\notag\\
=&\Dmb_{0,q}(t)+\frac{2\tau^2}{n}\w(t,0,q-1)+C'\eta
\end{align}
for a constant $C'$.  Therefore, by (\ref{eq:mb-v-t-k-q-v-t-k-q-1}), we
obtain
\begin{align}
\E_{\U_{t,k,q}}\left[\w(t,k,q)\right]
&=\w(t,k,q-1)-\eta\left[\Dmb_{0,q}(t)+\frac{2\tau^2}{n}\w(t,0,q-1)\right]+C'\eta^2.
\end{align}
By ignoring the term including $\eta^2$, 
we obtain
\begin{align}
\E_{\U_{t,k,q}}\left[\w(t,k,q)\right]&\simeq\w(t,k,q-1)-\eta
\left[\Dmb_{0,q}(t)+\frac{2\tau^2}{n}\w(t,0,q-1)\right]
\end{align}
for $k\ge 1$. The important point of this equation is that the update
term is not related to $k$ and it is calculated for the original data that is
the case of $k=0$ in DA. Note that we have
\begin{align}
\label{eq:v-t-0-Q-v-t-0-0}
\w(t,0,Q)&=\w(t,0,0)-\eta\Dmb_{0,1:Q}(t)
=\w(t-1)-\eta\Dmb_{0,1:Q}(t)
\end{align}
for $k=0$.
Therefore, by setting $(k,q)=(K,Q)$ and removing the symbol of the
expectation, we approximately obtain
\begin{align}
\label{eq:v-t-K-Q-v-t-K-0}
\w(t)&=\w(t,K,Q)\notag\\
&\simeq\w(t,K,0)
-\eta\sum_{q=1}^Q\left[\Dmb_{0,q}(t)+\frac{2\tau^2}{n}\w(t,0,q-1)\right]\notag\\
&=\w(t,K-1,Q)
-\eta\left[\Dmb_{0,1:Q}(t)+\sum_{q=1}^Q\frac{2\tau^2}{n}\w(t,0,q-1)\right]\notag\\
&\simeq\w(t,K-1,0)
-2\eta\left[\Dmb_{0,1:Q}(t)+\sum_{q=1}^Q\frac{2\tau^2}{n}\w(t,0,q-1)\right]\notag\\
&\cdots\notag\\
&=\w(t,1,0)
-K\eta\left[\Dmb_{0,1:Q}(t)+\sum_{q=1}^Q\frac{2\tau^2}{n}\w(t,0,q-1)\right]\notag\\
&\simeq\w(t,0,0)
-\eta\left[(K+1)\Dmb_{0,1:Q}(t)
+K\sum_{q=1}^Q\frac{2\tau^2}{n}\w(t,0,q-1)\right]\notag\\
&=\w(t-1)
-(K+1)\eta\left[\Dmb_{0,1:Q}(t)
+\frac{K}{K+1}\sum_{q=1}^Q\frac{2\tau^2}{n}\w(t,0,q-1)\right],
\end{align}
where the $6$th line comes from $\U_{t,0}=\O_{n,m}$.  This implies
that the update in the MB training under DA
is equivalent to obtaining $\w(t)=\w(t,Q)$ by
\begin{align}
\w(t,q)=\w(t,q-1)-\eta(K+1)\left[\Dmb_{0,q}(t)+\frac{K}{K+1}
\frac{2\tau^2}{n}\w(t,q-1)\right],
\end{align}
where we set $\w(t,0)=\w(t-1)$. Actually, it is easy to see that this
leads to (\ref{eq:v-t-K-Q-v-t-K-0}). This is the $\ell_2$ regularization
training under $\Smb_q(\w)+\lambda\|\w\|^2$ for the original data, in
which the regularization parameter is $\lambda=K\tau^2/(K+1)n$. However,
the learning rate is $(K+1)\eta$ which is larger than that in the naive
training and increases as $K$ increases. Hence, apparent acceleration
can be found also in the MB training.

We should mention the fact that this simple result is obtained by ignoring
the term including $\eta^2$.  This is true if $\eta$ is very small.
Additionally, since the term including $\eta^2$ is a function of the
update terms, this approximation can be precise when the amount of the
update is small. Obviously, such a situation occurs in the later stage
of training if the weight vector normally converges. Note that if we
turn our attention to neural networks, small amount of update of weights
are suggested; e.g., \cite{NTK}.

On the other hand, the MB training updates the weight vector several
times at each epoch.  Therefore, it seems to be capable of accelerating the
training. However, it is not straightforward since the update is made
for a part of data, by which it includes a kind of stochastic
fluctuation.  Nevertheless, our result supports this intuition with a
simple and explicit form that is approximately derived under a
stochastic evaluation.

\section{Numerical experiments}

\subsection{Example of linear regression}

\begin{figure}[ht]
\begin{center}
\begin{minipage}{50mm}
\begin{center}
\includegraphics[width=50mm]{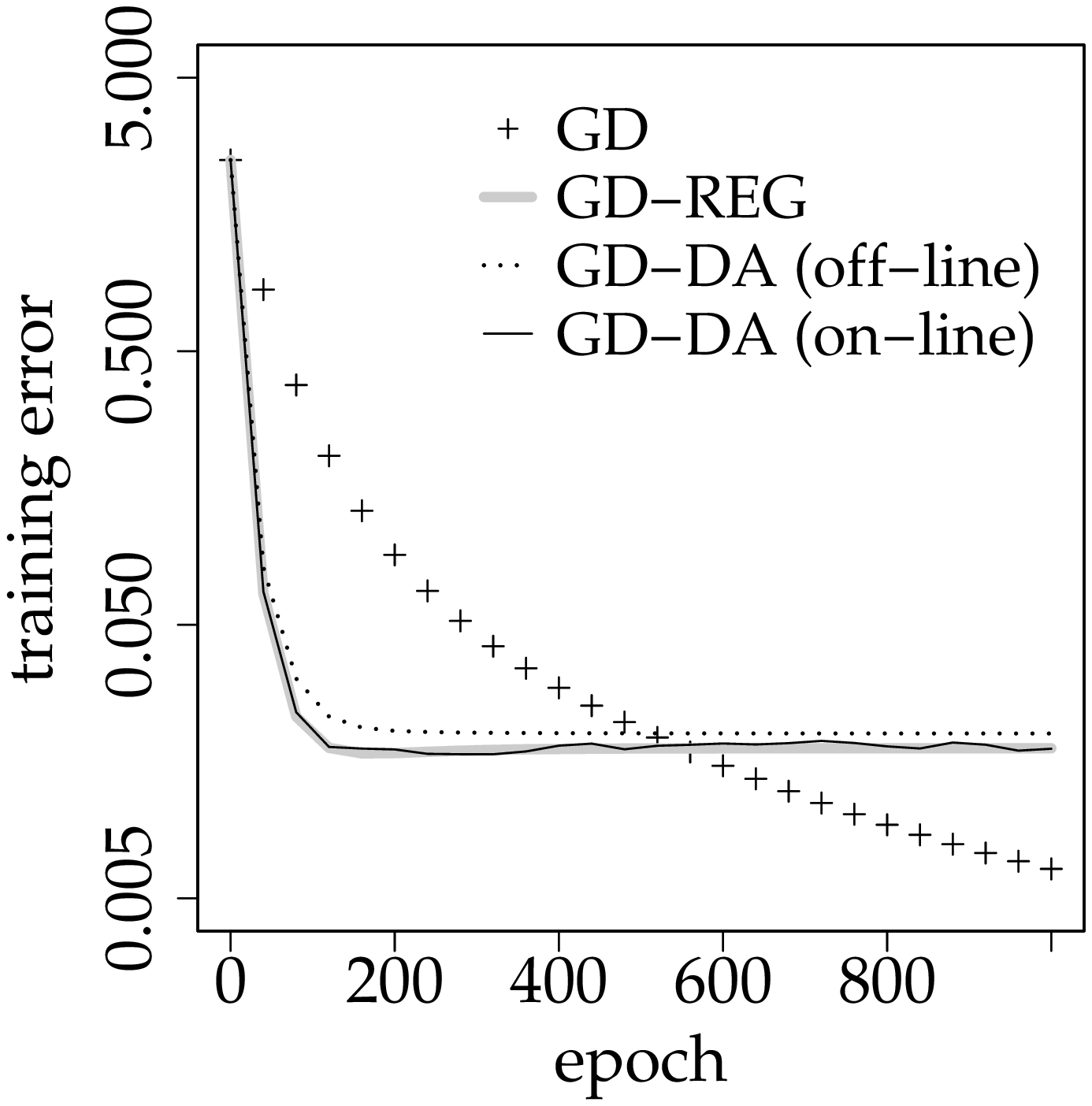}

(a) SSE training
\end{center}
\end{minipage}
\begin{minipage}{50mm}
\begin{center}
\includegraphics[width=50mm]{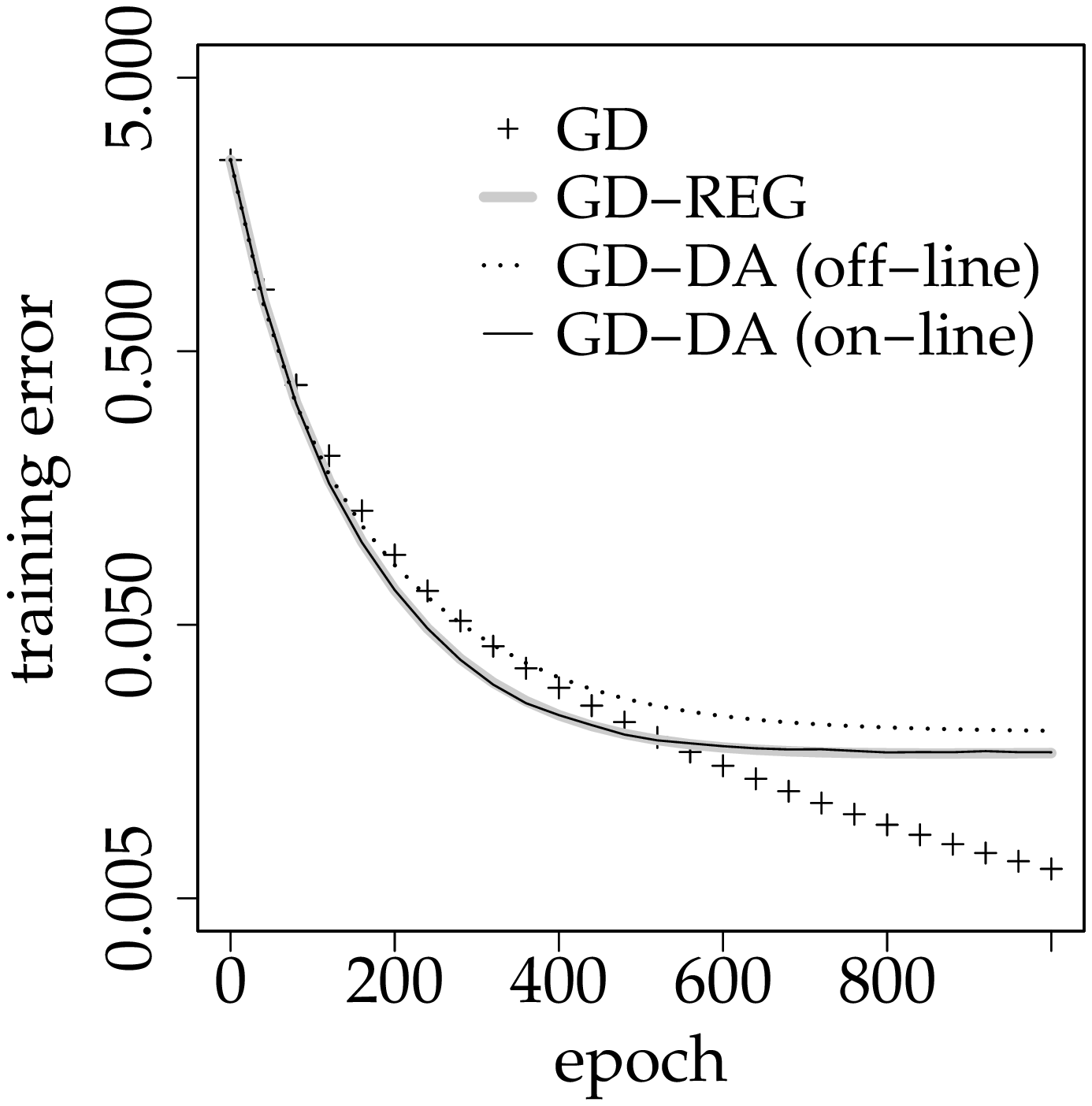}

(b) MSE training
\end{center}
\end{minipage}

\begin{minipage}{50mm}
\begin{center}
\includegraphics[width=50mm]{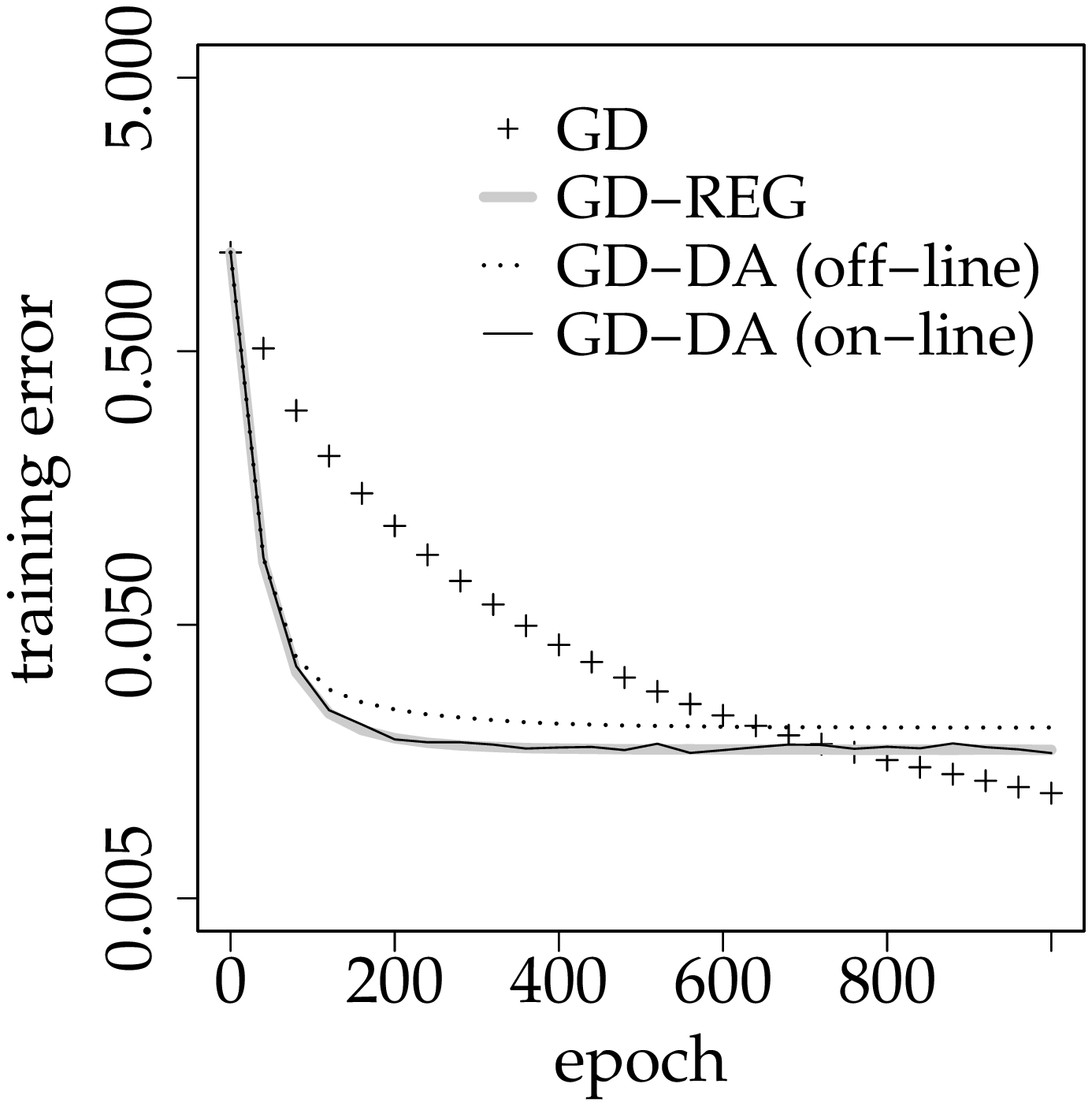}

(c) MB training ($K=4$)
\end{center}
\end{minipage}
\begin{minipage}{50mm}
\begin{center}
\includegraphics[width=50mm]{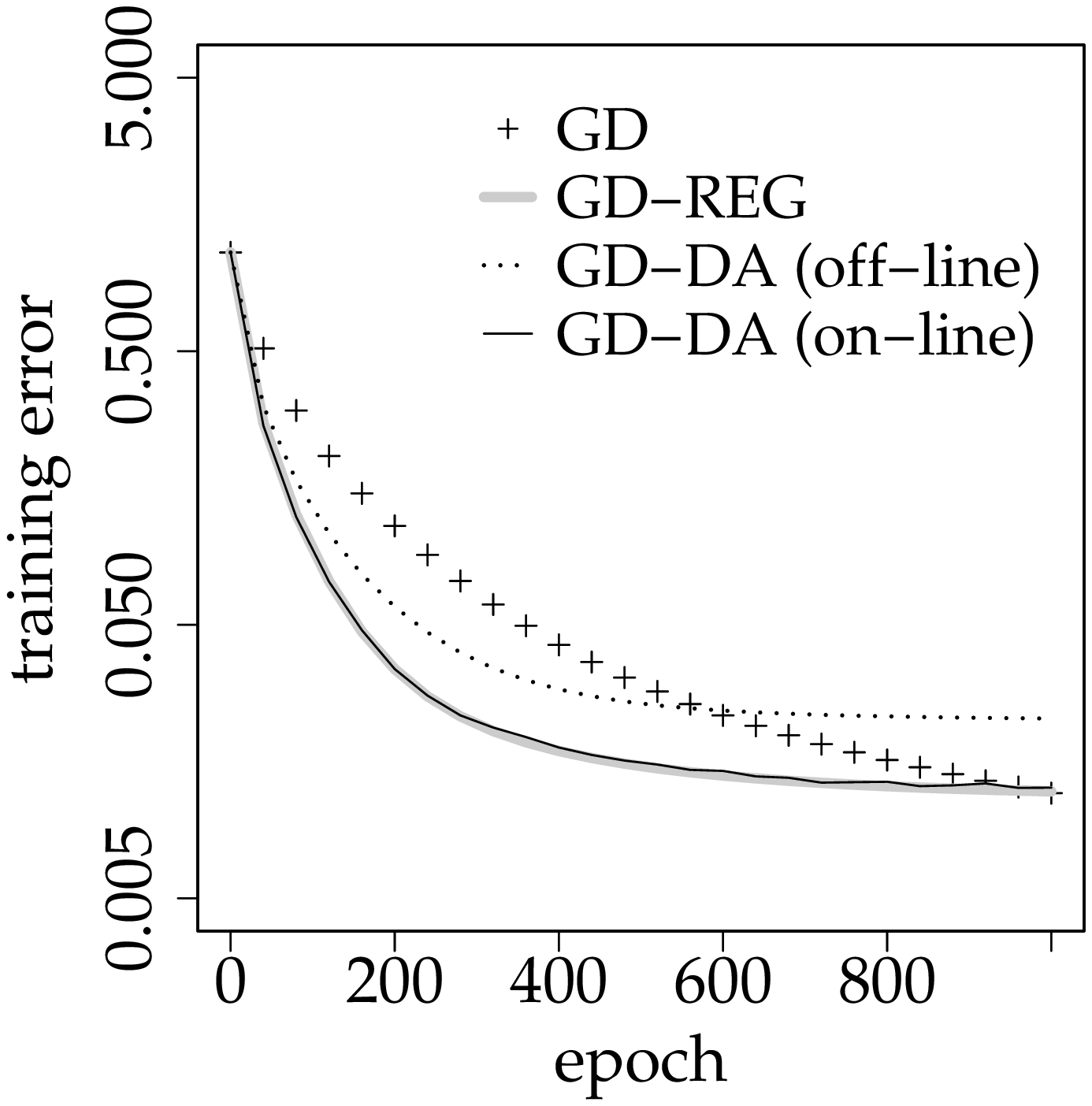}

(d) MB training ($K=1$)
\end{center}
\end{minipage}

\end{center}
\caption{Training curves for naive GD (GD), GD under the $\ell_2$
regularization (GD-REG ), GD under DA with off-line noise (GD-DA
(off-line)) and GD under DA with on-line noise (GD-DA (on-line)). (a)
SSE training, (b) MSE training, (c) MB training ($K=4$), (d) MB training ($K=1$).}
\label{fig:lr} 
\end{figure}

We consider a multiple regression. $x_{i,j},~i=1,\ldots,n,~j=1,\ldots,m$
are sampled from $N(0,\sigma_x^2)$ and
$y_i=x_{i,1}-x_{i,2}+\varepsilon_i$, where $\varepsilon_i\sim
N(0,\sigma^2)$.  The number of original data and input variables are set
to $n=20$ and $m=15$ respectively. This is an under-parametrization
scenario while $n$ is relatively small.  For DA, we set $Q=4$, thus
$\rho=5$. In DA, $u_{t,k}^{i,j}$ is sampled from $N(0,\tau^2)$.  Note
that any sampling here is at random.  We set $\sigma_x=0.5$,
$\sigma=0.2$ and $\tau=1.0$.  The training curves on the original
training data under the SSE , MSE and MB based training are shown in
Fig. \ref{fig:lr}.  The number of epochs is $1000$. The number of noisy
copies is $K=4$ for (a), (b) and (c).  For the MB training, we show the
result of $K=1$ in (d).  If the two weight updates are the same then
their training errors on original data are the same.  Therefore, in
Fig. \ref{fig:lr}, we show the transition of the MSE on the original
data.  In each figure, we show the training curves of the naive GD (GD
in the figure), GD with the $\ell_2$ regularization (GD-REG), GD under
DA with off-line noisy copies (GD-DA (off-line)) and GD under DA with
on-line noisy copies (GD-DA (on-line)).  For normalization, we set the
learning rate $\eta=0.001$ in the SSE training, $\eta=0.001n$ in the MSE
training and $\eta=0.001Q$ the MB training, where $Q$ is the number of
mini-batches. By this normalization, for example, the training curves
are consistent in (a) and (b) for naive GD. In the regularization
method, the values of the learning rate and regularization parameter are
set to the values derived in this paper. For example, in the MB
training, the learning rate is $(K+1)\eta$ and
$\lambda=K\tau^2/(K+1)n$.

In Fig. \ref{fig:lr}, we can observe the following facts.
\begin{itemize}
\item In all cases, the training curve under DA with on-line noisy copies is
      well consistent with that under the $\ell_2$ regularization
      training except a small fluctuation. This result supports our
      probabilistic analysis.

\item In all cases, the training curve of GD under DA with off-line
      noisy copies is different from that under the $\ell_2$
      regularization training. This is because the noise and updated
      weight vector are correlated as explained in the previous
      section. The difference is notable at the later stage of training
      while it is not observed in the early steps of training. This is
      because the correlation is small in this portion; i.e., in an
      extreme case, at the initial state, they are uncorrelated. And, it
      becomes larger as epoch increases. Therefore, the degree of the
      regularization is different for on-line and off-line noise. This
      determines the error in the later stage of training.

\item In (a), (c) and (d), the training curves of GD under DA (both of
      on-line and off-line) show the steep reduction in the early steps
      of training. This is caused by a large learning rates that is
      brought about by the number of copies in DA. It is not observed in
      (b); i.e., under MSE. Of course, for GD under DA, the error
      reduction is small at the later stage in training. This is caused
      by the regularization which avoids the over-training. The degree
      of over-fitting is controlled by the variance of noise in DA since
      it exactly corresponds to the regularization parameter.

\item By comparing (c) and (d), we can verify that the increase of the
      number of copies brings us apparent acceleration which is caused
      by the increase of the learning rate and contributes the early
      stage of training.  On the other hand, in this example, $K/(K+1)$
      is $0.8$ for $K=4$ and $0.5$ for $K=1$. Therefore, the
      corresponding regularization parameter is relatively different and
      this causes a difference in the training errors at the later
      stage. This is negligible when $K$ is large.

\end{itemize}

\begin{figure}[bt]
\begin{center}
\begin{minipage}{50mm}
\begin{center}
\includegraphics[width=50mm]{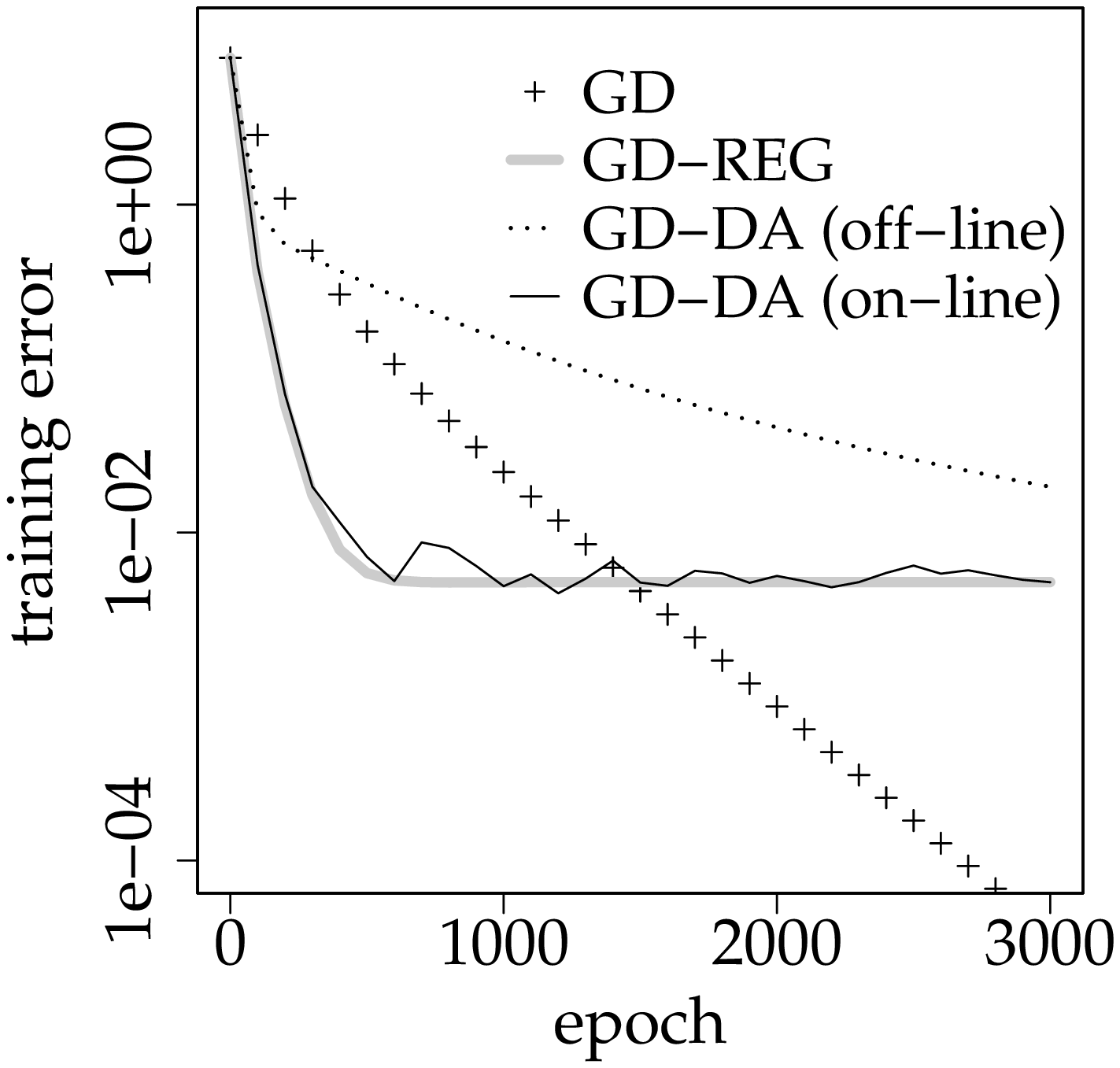}

(a) $K=2$
\end{center}
\end{minipage}
\begin{minipage}{50mm}
\begin{center}
\includegraphics[width=50mm]{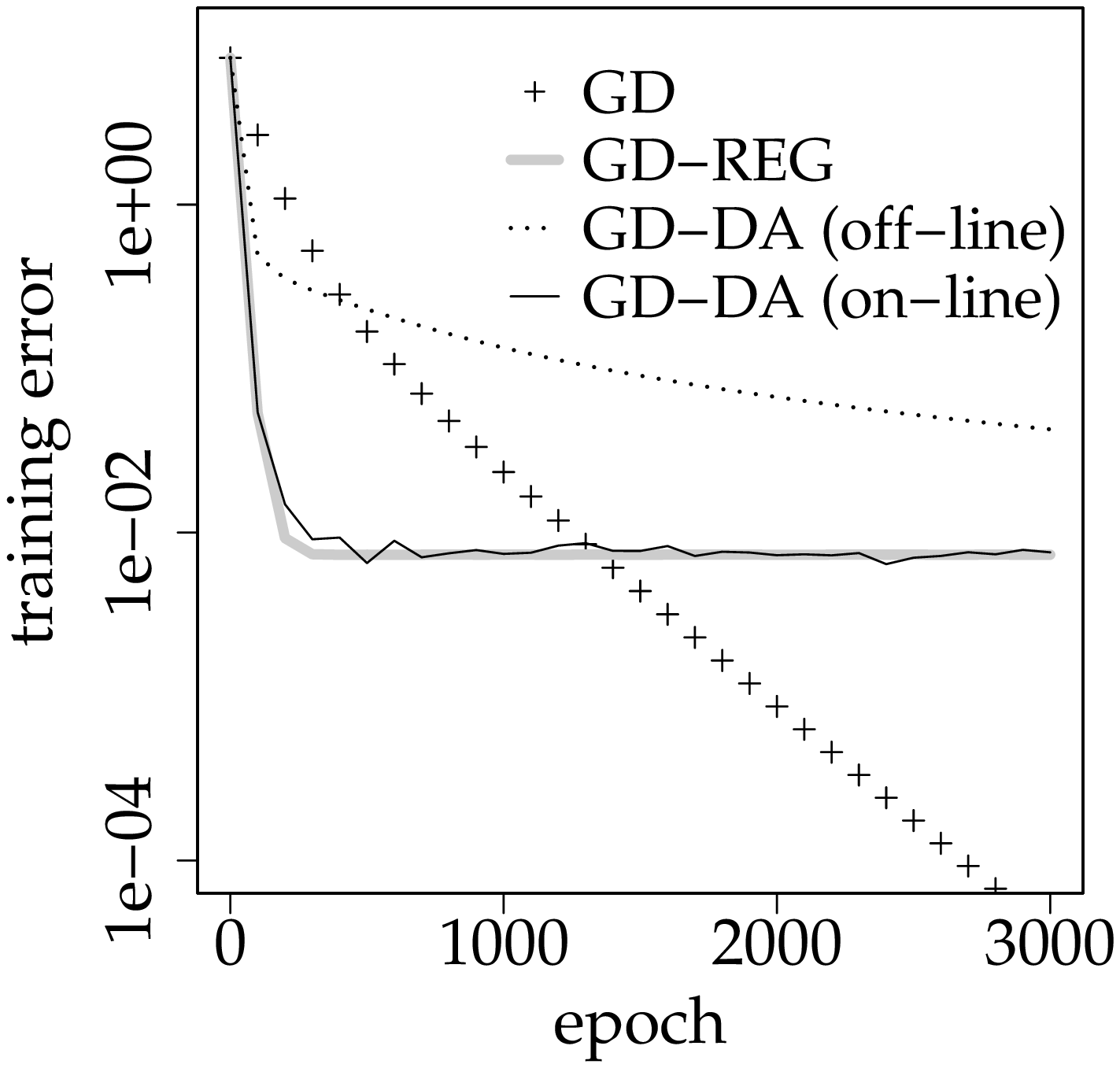}

(b) $K=5$
\end{center}
\end{minipage}

\end{center}
\caption{Training curves in an over-parametrization scenario. (a)
$K=2$, (b) $K=5$. }
\label{fig:lrop} 
\end{figure}

Our calculation can be applied regardless of the under-parametrized or
over-parametrized cases for original data; i.e., $n>m$ or $n<m$.  We
show the result of the MB training in the over-parametrized case.  The
setting of the numerical experiment is the same as above except that the
learning rate is $0.0001Q$ (it is smaller), the number of epochs is
$3000$, $\tau=2$ and $m=100$, in which we append $85$ irrelevant input
variables to the previous input.  Since $n=20<m$, this is an
over-parametrized case.  Note that a true relation is included in
assumed models here.  We show the training curves for the cases of $K=2$
and $K=5$ in (a) and (b) of Figure. \ref{fig:lrop} respectively.  The
number of whole data (data including nosy copies) is $3n=60$ for the
former and $6n=120$ for the latter.  Therefore, under the DA, the former
is the over-parametrized case and the latter is the under-parametrized
case. As seen in the figure, the training curves of GD-DA (on-line) for
both cases are almost consistent. Since adding on-line noisy copies is
equivalent to the $\ell_2$ regularization, DA with noisy copies may be
effective for avoiding over-training regardless of the
under-parametrization or over-parameterization scenarios. However, the
increase in the total number of data by DA may not so important other
than the apparent acceleration of training.  On the other hand, the
regularization effects for both type of DA are very similar in
under-parametrization scenario.  However, they are not in
over-parametrization scenario. This is because, in the over-
parameterization scenario, over-ﬁtting to noise injected before training
is serious; i.e., the training process depends largely on the injected
noise.

\subsection{Example of neural networks}

As pointed out in recent works, wide neural networks can be linearized
under the GD training. Therefore, our result on linear regression is
possible to be applied to neural networks. We test this through simple
numerical examples for real datasets.

\begin{figure}[t]
\begin{center}
\begin{minipage}{50mm}
\begin{center}
\includegraphics[width=50mm]{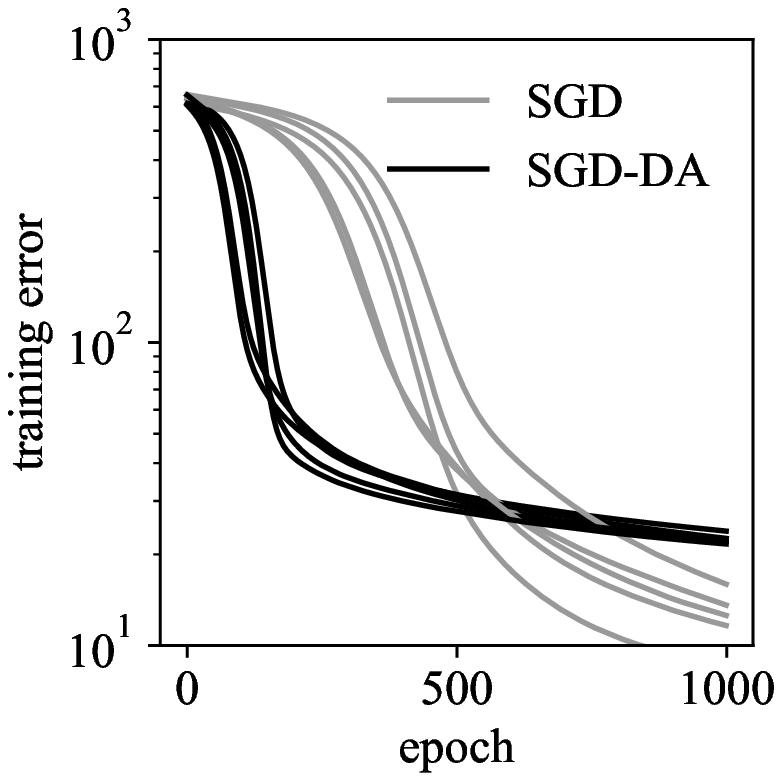}

(a) Auto-MPG (BS is $20$)
\end{center}
\end{minipage}
\begin{minipage}{50mm}
\begin{center}
\includegraphics[width=50mm]{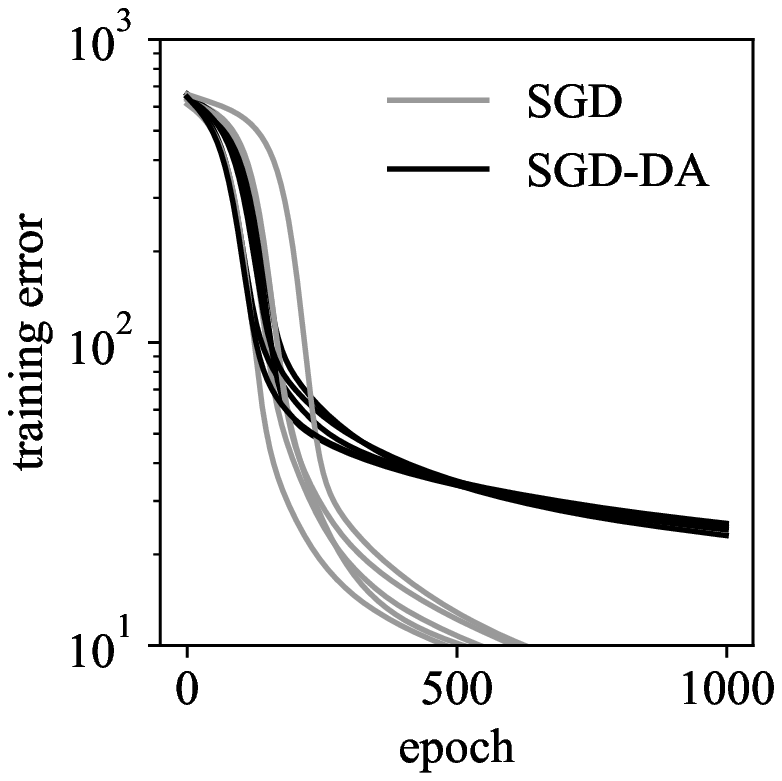}

(b) Auto-MPG (FB)
\end{center}
\end{minipage}

\begin{minipage}{50mm}
\begin{center}
\includegraphics[width=50mm]{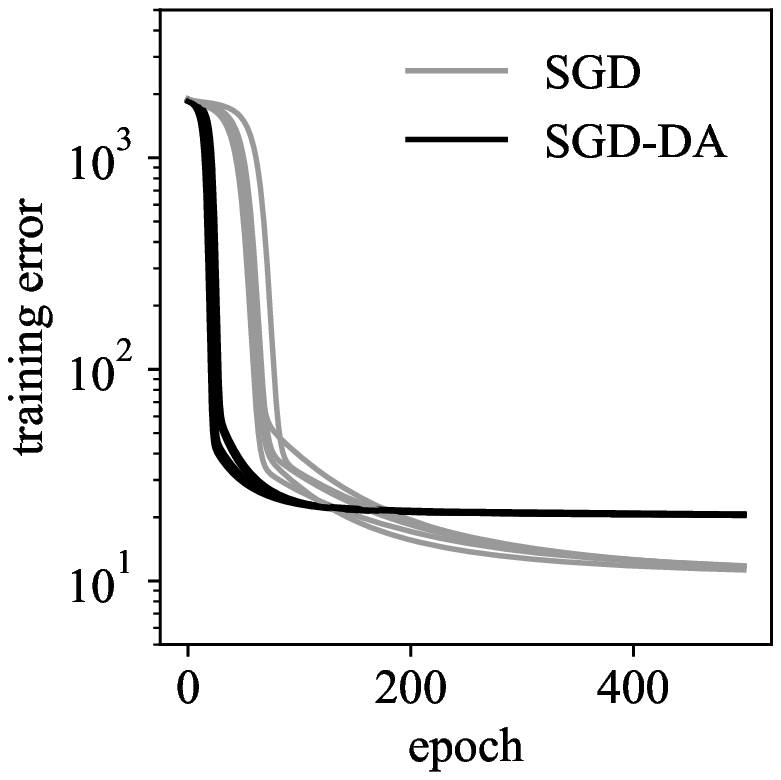}

(c) SIEC (BS is $100$)
\end{center}
\end{minipage}
\begin{minipage}{50mm}
\begin{center}
\includegraphics[width=50mm]{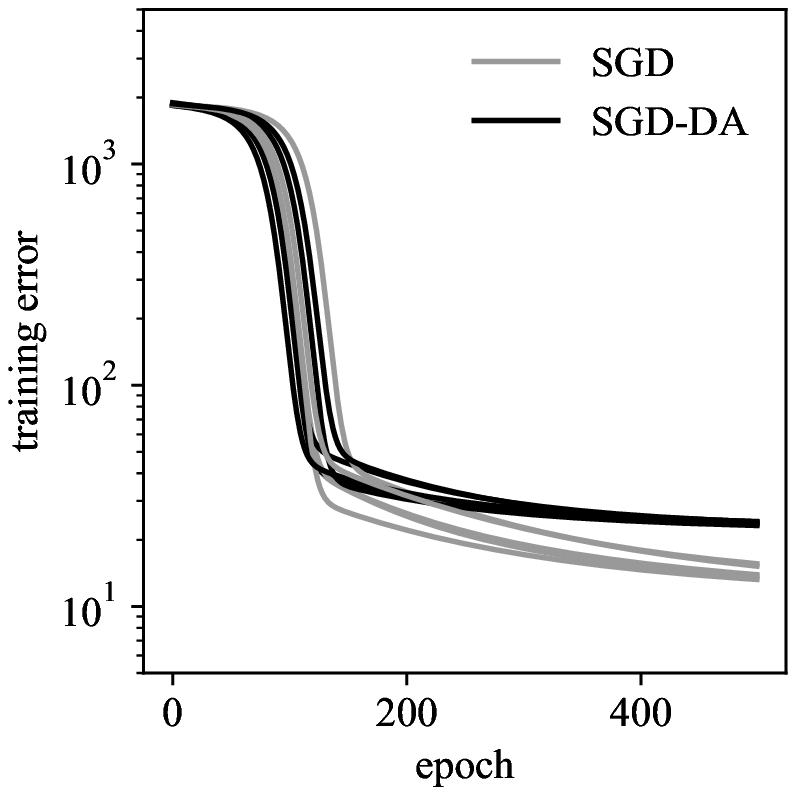}

(d) SIEC (FB)
\end{center}
\end{minipage}

\end{center}
\caption{Training curves for a naive SGD (referred to as SGD) and SGD
with off-line DA (referred to as SGD-DA) under different datasets and batch
sizes (BS).}
\label{fig:nn-test} 
\end{figure}

We use the two benchmark datasets from UCI repository\cite{UCI}, which
are the Auto-MPG (miles per gallon) and SIEC (steel industry energy
consumption) datasets.  The number of data is $78$ for Auto-MPG and
$1752$ SIEC which are $20[\%]$ and $5[\%]$ of the available data
respectively. The architecture of neural network is $4$-layer for both
datasets, in which the number of nodes at each layer is 8-32-32-1 for
Auto-MPG and 6-64-64-1 for SIEC. For both datasets, the activation
function is ReLU in the two hidden layers and linear in the output
layer. We use a simple Stochastic GD (SGD) training which is implemented
in Keras\cite{Keras} since our purpose is to observe the property of the
GD training. The error function is the MSE with different batch size.
The learning rates for both datasets are set to $0.00001$ in the MB
training and $0.0001$ in the FB training.  The number of epochs are set
to $1000$ for Auto-MPG and $500$ for SIEC. In Fig.\ref{fig:nn-test}, we
show the MSE based training curves for $5$ runs of a naive SGD (referred
to as SGD) and SGD under DA (referred to as SGD-DA) which is off-line DA
with $K=2$. $\tau$ is set to $0.2$. Note that the training curves are
the MSE errors for the training data; i.e., original data for SGD and
augmented data for SGD-DA.  We show the results of the MB and FB
training. The batch size (BS) in the MB training are $20$ for Auto-MPG
and $100$ for SIEC respectively.  Therefore, the results in (a) and (c)
correspond to the MB training and those in (b) and (d) correspond to the
FB training under the MSE criterion in our analysis. We summarize the
result below.

\begin{itemize}
 \item As seen in (b) and (d) which are the FB settings, the rate of
       error reduction at the initial portion of training is the same
       for both cases; i.e., SGD and SGD-DA. On the other hand, the
       training is suppressed in SGD-DA.  This is caused by the
       regularization effect by DA. Therefore, apparent acceleration of
       DA can not be observed under the setting of the FB training under
       the MSE criterion. This is consistent with our result on linear
       regression.

\item As seen in (a) and (c) which are the MB settings, the
       regularization effect by DA can be observed. Additionally, the
       steep reduction of error at the initial portion of training is
       observed for SGD-DA. Thus, apparent acceleration caused by
       introducing DA is observed. This is also consistent with our
       result on linear regression.

\end{itemize}
In case of the MB training of linear regression, DA using noisy copies
brings us the effects of regularization and apparent acceleration of
training. The latter may not essential in linear regression problems
since there is only one global minimum. However, in nonlinear regression
problems including layered neural networks, it affects the convergence
property of training since the error surface of neural networks is
complex.

\section{Conclusions and future works}

We analyzed the training process of the GD update for linear regression
under DA with on-line noisy copies, in which noise of copy is injected
into input.  We especially evaluated the average behavior of the
training process of the FB training under SSE, MSE and the MB training
under MSE.  This result was verified in a simple numerical experiment.
We also experimentally investigated the effect of DA on neural
networks. This numerical experiment suggests that our analysis on linear
regression is possible to apply to neural networks. As a future work, we
need to analyze the effect of DA with noisy copies on the training
process of neural networks. The analysis on the regularization effect of
layered and convolutional neural networks may be followed by
\cite{Dhifallah2021}. On the other hand, it is worthwhile to note that a
mixture of apparent acceleration and regularization effect obtained by
DA has a possibility of different effect on training process of neural
networks. This is because there are bottlenecks in training of neural
networks; e.g. \cite{Dinh2017}. Because of the complex error surface in
neural networks, there is a possibility of the existence of the effect
that is not appeared in the analysis of linear regression; i.e., the
other than a simple apparent acceleration and regularization. We need to
clarify it in the future work.

\section*{Acknowledgment}

This work was supported by Japan Society for the Promotion of Science
(JSPS) KAKENHI Grant Number 21K12048. The author thanks Kenji Nawa and
Kohji Nakamura in Faculty of Engineering, Mie University for their
advices.

\end{document}